\documentclass{article}
\usepackage{ijcai19}

\usepackage{amsfonts}
\usepackage{times}
\usepackage{soul}
\usepackage{url}
\usepackage[hidelinks]{hyperref}
\usepackage[utf8]{inputenc}
\usepackage[small]{caption}
\usepackage{graphicx}
\usepackage{amsmath}
\usepackage{booktabs}
\usepackage{algorithm}
\usepackage{algorithmic}
\usepackage{bm}

\title{Interpretable Text Classification Using CNN and Max-pooling}

\author{
  Hao Cheng$^1$\and
  Xiaoqing Yang$^1$\footnote{Contact Author}\and
  Zang Li$^1$\And
  Yanghua Xiao$^2$\\
  Yucheng Lin$^1$\\
  \affiliations
  $^1$diegochenghao\_i\\
  $^1$yangxiaoqing\\
  $^1$lizang\\
  $^2$shawyh\\
  $^1$linyucheng\\
  \emails
  first@didiglobal.com,
  second@fudan.edu.cn
}

\begin{document}

\maketitle

\begin{abstract}
	 Deep neural networks have been widely used in text classification. However, it is hard to interpret the neural models due to the complicate mechanisms. In this work, we study the interpretability of a variant of the typical text classification model which is based on convolutional operation and max-pooling layer. Two mechanisms:\emph{convolution attribution} and \emph{n-gram feature analysis} are proposed to analyse the process procedure for the CNN model. The interpretability of the model is reflected by providing posterior interpretation for neural network predictions. Besides, a multi-sentence strategy is proposed to enable the model to beused in multi-sentence situation without loss of performance and interpretability. We evaluatethe performance of the model on several classification tasks and justify the interpretable performance with some case studies.
\end{abstract}

\section{Introduction}
In the field of natural language processing, text classification is an important task with a lot of downstream applications such as sentiment analysis \cite{go2009twitter}, relation classification \cite{Lin2016NeuralRE}, and document classification \cite{yang2016hierarchical}, etc.
Most of the research on text classification focus on how to extract effective features from text and construct accurate classifiers to integrate the features \cite{mccallum1998comparison}. In deep neural methods, word vectors project the semantic information of words into a dense low-dimension space where the semantic similarity of words can be assessed by Euclidean distance or cosine similarity. Deep learning methods performing composition over word vectors to extract features, has been proven to be effective classifiers and achieve excellent performance on different NLP tasks (\cite{Lin2016NeuralRE,Kim2014ConvolutionalNN,liu2016recurrent}).
 
Among the text classification deep neural methods, Recurrent Neural Network (RNN) and Convolutional Neural Network (CNN) have been widely used in a lot of applications. RNN can capture dependencies between different components in a statement. 
CNN, employing convolution operation (\cite{LeCun1998GradientBasedLA}) to capture features, is originally used in computer vision. 
A plenty of work adopt the CNN network structure for the NLP tasks (\cite{Lin2016NeuralRE,Kim2014ConvolutionalNN}). The results prove that CNN has a unique advantage in capturing local features of word vectors. 

However, the deep neural methods still have a severe challenge: it is hard to interpret the classification results. For an end-to-end deep neural architecture, the intermediate hidden states are usually some real-valued vectors or matrices which are difficult to figure out the semantic meaning or to establish an association with target categories. 
Recent work on interpretability has focused on producing posteriori explanations for neural network predictions~\cite{alvarez2018towards}. There has been some related research on neural networks based RNN~\cite{ding2017visualizing,karpathy2015visualizing,hiebert2018interpreting}.  However, interpreting CNN-based models remainsan under-explored area. There have been plenty of research and applications interpreting and visualizing CNN in the field of computer vision~\cite{zeiler2014visualizing,zeiler2011adaptive,krizhevsky2012imagenet}. However, due to the different for msof input data, it is hard to apply the methods designed for picture pixels on the word vectors. \cite{goldberg2016primer} studies how does CNN based classifiers works. The procedure can be divided into two steps: (1) 1-dimensional convolution operations are used to detect closely-related familyof n-grams. (2) Max-pooling interprets the n-grams features, which is further used to make the decision using linear layers. In both parts, each unit of the output have complicate association with multiple inputs units, and the non-linear layers and the complex structure of the convolutional layer exacerbate the difficulty of interpretation. \cite{jacovi2018understanding} interprets CNN models by extracting and analyzing the n-gram features the modelhas learned. But their research does not establish an association between input tokens and n-grams features.

In this paper, we implement the interpretabilityof a variant of typical CNN based classifier. For each process procedure of the CNN based model,we propose a method to establish the point wise associations between the input and the output respectively. The max-pooling layer is used to interpret all the n-gram features to generate the sentence representations. In our work, the interpreting operations of the proposed model consists of two parts: \emph{convolution attribution} and \emph{n-gram feature parsing}. Convolution attribution involves the disassembly of the convolution operation. We construct the association between n-gram features and word vectors by analyzing the contribution of the values in each word vector to the convolution results. N-gram feature parsing constructs an association between n-gram features with the classification result by backtracking pooling operations. Besides, a multi-sentence strategy is proposed for the situation where the input text contains multiple sentences.

Common wisdom suggests that interpretabilityand performance currently stand in apparentconflict in machine learning \cite{alvarez2018towards,mori2019balancing}.  In experiments, we comparethe classification performance of the classificationmodel with several state-of-the-art classifiers firstly.Besides, we validate the interpretable performanceof the model visualize the predictions in 2scenarios, namely \emph{interpreting predictions with input tokens} and \emph{interpreting predictions with samples}.

\section{Related Works}

In this section, we review the related studies in two aspects, respectively text classification and neural network interpretable.

Among the large number of recent works employing deep neural networks on text classification task, our model use a quite common structure and has similarities to many models. 
Compared with \emph{textCNN} in \cite{Kim2014ConvolutionalNN}, we employ pooling operations $n$-grams other than the vector dimensions. Our model is also similar to the model proposed in \cite{Lin2016NeuralRE} The difference is that we employ multiple size of convolution kernels to capture features. There are some other variants similar to the model, such as \cite{zhang2015character}. 

The research on interpreting NLP neural networks involve various directions. \emph{Layer-wise relevance propagation} (LRP) \cite{bach2015pixel} is widely used in plenty of interpretable researches, our methods also refer to this idea. LRP backpropagating relevance recursively from the output layer to the input layer, is originally designed to compute the contributions of single pixels to predictions for image classifiers. \cite{ding2017visualizing} uses LRP to interpret a attention-based encoderdecoder framework for neural machine translation. \cite{croce2018explaining} explain neural classifiers by providing the examples which motivate the decision. They employ LRP to trace back the model predictions and further provide human-readable justification. \emph{Attention mechanism} \cite{bahdanau2014neural} is used in various NLP situations to adjust the weight of different parts of the input. Some research analyze and visualize the weights of the attention mechanism to provide humanreadable explanations for the prediction \cite{yang2016hierarchical,luong2015effective}.  Some researches propose new neural network architectures that are easier to interpret. \cite{stahlberg2018operation} propose a neural machine translation model that incorporate explicit word alignment information in the representation of the target sentence, the interpretability is implemented by generating a target sentence in parallel with the source sentence. Although the definition of interpretability differs in various situation, most of related research indicate that where is conflict between the interpretability and performance for machine learning \cite{mori2019balancing,murdoch2019interpretable,rudin2018please}. New neural architectures are usually proposed to take both requirements into account \cite{alvarez2018towards}.

\section{Methods}

\begin{figure}
	\centering
	\includegraphics[width=.45\textwidth]{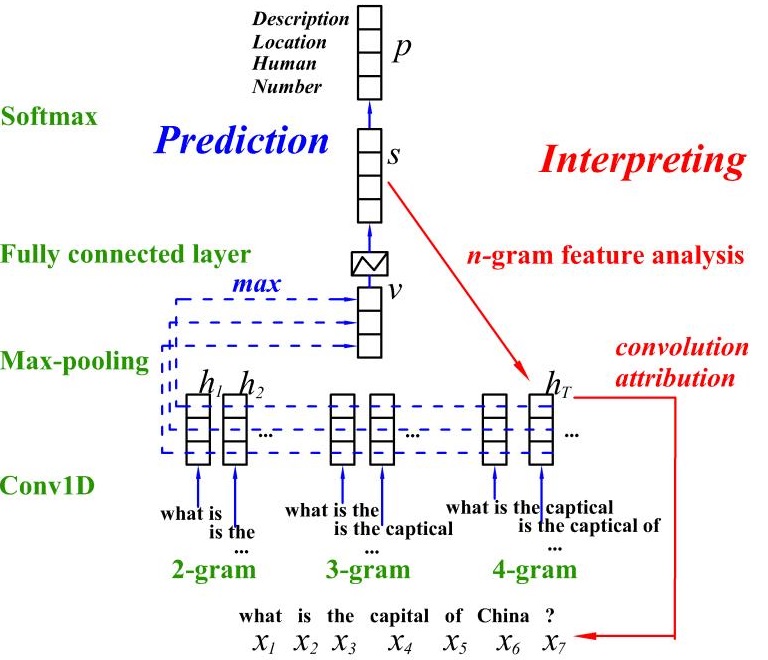}
	\caption{The framework of the proposed text classification model. The red lines indicate the process of generating features and making the predictions. The blue lines indicates the direction of the interpreting strategies in our model.}
	\label{framework}
\end{figure}

In this section, we introduce the structure of our proposed text classification model and the methods to interpret the classification result. The framework of the proposed model is shown in the Figure \ref{framework}.   

\subsection{Text Classification Model}
As shown in Figure \ref{framework}, the input of the text classification model is a sequence of $w$ words. We convert each word into a word representation vector with $d_{e}$ dimensions, denoted as $X=\{\bm{x}_1, \bm{x}_2, ..., \bm{x}_{w}\}$ ($\bm{x} \in \mathbb{R}^{d_{e}} $). We employ convolution operations with different kernel size ($2, 3, ..., l$, $l$ denotes the maximum convolution size) to derive $n$-grams. These $n$-grams serve as the feature for classification. 
The number of the convolution filters decides the dimension of feature vectors. Which is set to $d_m$ for all kernel size of convolution operations.
We denote all the $n$-gram features as $\bm{h}_{1:T} \in \mathbb{R}^{d_m}$, where $T$ is the total $n$-gram number, which is related with $w$ and $l$.  Actually, the text classification tasks often face the challenge that the length of the input sentence is a variable. In our model, max-pooling layer is used to integrate the $n$-gram features and generate a sentence vector $\bm{v}$ with the fixed dimension $d_m$. The sentence vector $\bm{v}$ is then fed into a fully-connected layer to generate a score distribution $\bm{s} \in \mathbb{R}^{n_{t}}$ over $n_{t}$ target categories. The final output of the model $\bm{p} \in \mathbb{R}^{n_{t}}$ is a probability distribution on target categories.

The convolution operations used in our methods, are the same with the architecture of \cite{collobert2011natural}. Generally, we represent the $n$-gram $\bm{x}_{i:i+n-1}$ as the concatenation of words $\bm{x}_i, \bm{x}_{i+1}, ..., \bm{x}_{i+n-1}$. Suppose the feature vector $\bm{h}_t ( 1 \leq t \leq T)$ is related with $\bm{x}_{i:i+n-1}$, $\bm{h}_t$ is calculated by: 


  \begin{equation}
  \bm{h}_t = W_n \cdot x_{i:i+n-1}+\bm{b}_n
  \end{equation}
  where $W_n \in \mathbb{R}^{n*d_{e}*d_m}$ and $\bm{b}_n \in \mathbb{R}^{d_{m}}$ are parameters for convolution operation. The size of the convolution kernels is equal to the length of the $n$-grams.
  
  The filter is applied to every possible window in $X$. 
  The max-pooling operation captures the maximum value of the feature vectors in each dimension over the $n$-grams to generate the sentence vector $\bm{v}$, which is defined as:
   \begin{equation}
    \bm{v} = max[\bm{h}_{1:T}],	
   \end{equation}
 
  The sentence vector $\bm{v}$ is then connected to a fully connected layer to calculate the score distribution on target categories. The score distribution $\bm{s} \in \mathbb{R}^{n_{t}}$ and the probability distribution $\bm{p} \in \mathbb{R}^{n_{t}}$ is respectively calculated by:
  \begin{equation}
  	\bm{s} = M \cdot \bm{v} + \bm{b},
  \end{equation}
  \begin{equation}
   \bm{p}[i] = \frac{exp(\bm{s}[i])}{\sum_{j=1}^{n_{t}}{exp(\bm{s}[j])}},
  \end{equation}
  where $M$ and $\bm{b}$ are the parameters of the fully connected layer, $n_{t}$ is the number of target categories.

\subsection{Interpretable Principles}
\label{interpretable principles}
In this part, we introduce the methods to interpret the classification predictions of the proposed model. As shown in Figure \ref{framework}, essentially, the interpreting process of the model is the inverse derivation of the model predictions. The implementation of the interpretability consists of two parts: \emph{convolution attribution} and \emph{n-gram feature analysis}.

\subsection*{Convolution Attribution}
\begin{figure}
	\centering
	\includegraphics[width=.36\textwidth]{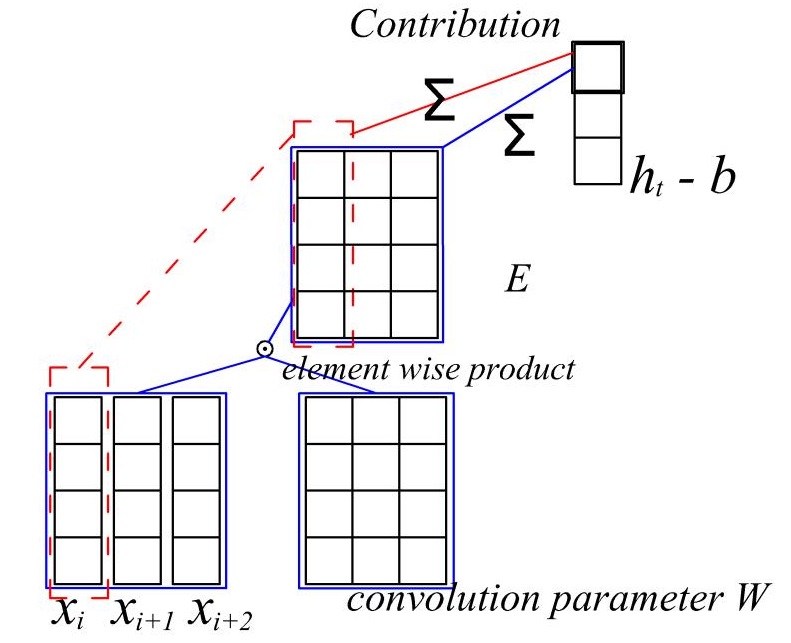}
	\caption{\emph{convolution attribution}. We construct the relevance between the word vector $\bm{x}$ and n-gram features $\bm{h}_t$ by analyzing the contributions of the values of word vector to each dimension of the convolution result.}
	\label{attribution} 
\end{figure}
In the proposed model, each feature vector obtained by the convolution operation correspond to a $n$-gram. In this part, we propose the method \emph{convolution attribution} to analyze each convolution operation independently and obtain the relevance between word vectors and the $n$-gram features. 


Let $\bm{h}_t \in \mathbb{R}^{d_{m}}$ denote the feature vector for the $n$-gram $\bm{x}_{i:i+n-1} \in \mathbb{R}^{n*d_{e}}$, $W \in \mathbb{R}^{n*d_{e}*d_{m}}$ and $\bm{b}_n \in \mathbb{R}^{d_m}$ are the parameters for the convolution layer. We define the result of the element-wise product for the $k$-th kernel as $E_k \in \mathbb{R}^{n*d_{e}}$:

\begin{equation}
E_k=\bm{x}_{i:i+n-1} \odot W[:,:,k]
\end{equation}
It is obvious that $\bm{h}_t$ is calculated by:
\begin{equation}
\bm{h}_t[k]=\sum_{i,j}{E_k{[i,j]}} + \bm{b}_n[k]
\end{equation}
where $\bm{b}_n$ is the bias of the convolution.
We define the contribution of word $\bm{x}_g$ in the n-gram $ \bm{x}_{i:i+l-1}$ ($i \leq g < i+n$) for the $k$-th dimension of the feature vector $\bm{h}_t$ as $ Rel_{\bm{x}_g \to \bm{h}_t[k]}$, which is calculated by:
\begin{equation}
Rel_{\bm{x}_g \to \bm{h}_t[k]} = 
\begin{cases}
\frac{\sum^{d_{e}-1}_{q=0}{E_k[g,q]}}{\bm{h}_t[k]-\bm{b}_n[k]} & if \left|\bm{h}_t[k]-\bm{b}_n[k]\right|>\epsilon\\
\frac{1}{n} & otherwise
\end{cases}
\end{equation}
if the sum of $E$ is too small, we simplify the contribution distribution on corresponding dimension. 

We take all the contributions for each word of $\bm{x}_{i:i+n-1}$ on every dimension of the feature vector $\bm{h}_t$ into account and build the matrix $Rel(\bm{h}_t) \in \mathbb{R}^{l*d_m}$. The sum of each column of the matrix $Rel(\bm{h}_t)$ is $1$, denoting the contribution distribution of the words $\bm{x}_{i:i+n-1}$ on a specific dimension of the feature vector $\bm{h}_t$.

\subsection*{N-gram Feature Analysis}
\begin{figure}
	\centering
	\includegraphics[width=.45\textwidth]{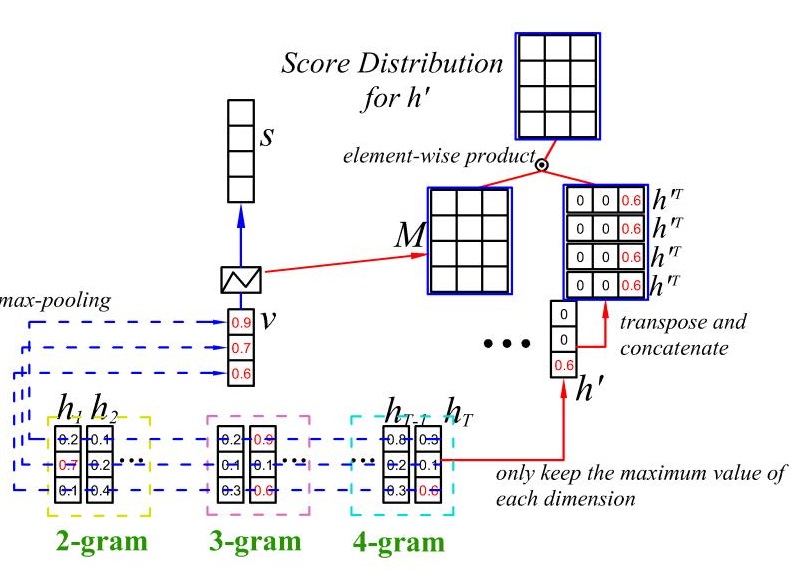}
	\caption{The structure of \emph{n-gram feature analysis}. \emph{Score distribution} denotes the effect of each dimension of the $n$-gram feature $\bm{h}'$ on target categories.}
	\label{n-gram feature analysis}
\end{figure}

\emph{N-gram Feature Analysis} splits the score distribution $\bm{s}$ corresponding to sentence vector $\bm{v}$ into the sum of the contributions of n-gram features $\bm{s}_{1:T}'$.
In the proposed classification model, all the feature vectors $\bm{h}_{1:T}$ and the sentence vector $\bm{v}$ share the same size. The sentence feature vector $\bm{v}$ integrates all the $n$-gram features with a max-pooling. As can be seen, the feature vectors also share the same structure. We analyze the effect of the feature vectors on the final classification prediction using the fully connected layer which is connected to sentence vector $\bm{v}$. The details are shown in Figure \ref{n-gram feature analysis}.

The analysis process is performed independently for each $n$-gram. First, we figure out which parts of each feature vector $\bm{h}$ actually play a role in the classification process using reverse derivation of the max-pooling layer. For each dimension of any feature vector, if the value is equal to the corresponding dimension of the sentence vector $\bm{v}$, or in other words, if the value in this dimension is the largest among all the n-gram features, we keep this value. Otherwise we set the value at this position as 0. We define the modified vector $\bm{h}'$ by:



  \begin{equation}
  \bm{h}'_{t}[j]=
  \begin{cases}
  \bm{h}_{t}[j] & \text{ if $\bm{h}_{t}[j]= max(\bm{h}_{1:T}[j])$}\\
  0 & \text{otherwise}
  \end{cases}
  \end{equation}

We pass the disassembled vector $h'_t$ through the fully connected layer in the original model to get a score distribution $s'_t$ on target categories. We only focus on the differences in the contributions of different features on categories. Therefore, the offset $b$ in the fully connected layer is ignored. The score distribution $s'_t \in \mathbb{R}^{n_t}$ is calculated by

\begin{equation}
  s'_t= M \cdot h'_t
  \label{s related 1} 
\end{equation}

If the values of all the features in a specific dimension are less than 0, $s'$ may differ from the real situation. This can be avoided by linking a \emph{Rectified Linear Unit (Relu)} activation function with the results of the convolution layers in the proposed model.

$\bm{s}'$ denotes the score distributions on target categories for the $n$-gram feature $\bm{h}'_t$. We further analyze $s'$ among the vector dimension of $h'_t$. The analysis process is implemented by converting the form of Equation \ref{s related 1}. We define the operations as following:

\begin{equation}
S'_t = Concatenate(\underbrace{{\bm{h}'_t}^{T}, {\bm{h}'_t}^{T}, ..., {\bm{h}'_t}^{T}}_{n_t})
\end{equation}
\begin{equation}
Sco(\bm{h}'_t) = S'_t \odot M
\label{score distribution}
\end{equation}
where $S'_t \in \mathbb{R}^{d_m*n_t}$ is the matrix obtained by repeating the transposition of $h'_t$ by $n_t$ times and concatenating by column axis. The matrix $S'_t$ has the same dimensions with the weighting matrix of the fully connected layer $M$. We define the element-wise product result between $S'_t$ and $M$ as $Sco(\bm{h}'_t)$. The values of $Sco(\bm{h}'_t)$ denote the score distributions for each dimension of feature vector $\bm{h}'_t$ on target categories.

\section*{Interpretable Features Aggregation}
We define the contribution matrix $Rel(\bm{h}_t)$ and the score distribution matrix $Sco(\bm{h}'_t)$ in \emph{convolution attribution} and \emph{n-gram feature analysis} respectively. The contribution matrix $Rel(\bm{h}_t)$
represent the contribution of word vectors on each dimension of the $n$-gram feature and the score distribution matrix $Sco(\bm{h}'_t)$ measures the impact of each dimension of $n$-gram feature on target categories. We integrate the two matrices to evaluate the impact of each word vector on the classification results.

In the $n$-gram feature $\bm{h}'_t$, the values on target categories for the word vector $\bm{x}_{i+q}$, $Val_{\bm{h}_t} (\bm{x}_{i+q})$, is calculated by:
\begin{equation}
Val_{\bm{h}_t} (\bm{x}_{i+q}) = (Rel(\bm{h}_t) \cdot Sco(\bm{h}'_t))[q,:]
\label{score distribution}
\end{equation}
The value of word vector $\bm{x}_i$ on target categories in the overall sentence $Val(\bm{x}_i)$ obtained by:

\begin{equation}
Val(\bm{x}_{i}) = \sum_{\bm{h}_g}^{J_i}{Val_{\bm{h}_g}(\bm{x}_i)}
\label{score distribution}
\end{equation}
where $J_i$ is the set of $n$-gram features where each of $n$-grams contains $\bm{x}_i$ inside.
Integrating the results of all the word vectors in $X$, we define value matrix $Val(X) \in \mathbb{R}^{n_w*n_t} $ as the final interpretable result. $Val(X)$ shows the value of each word vector on every target categories.

\subsection{Multi-sentence weighting strategy}
For input context with any length, our model outputs a sentence feature vector with a fixed size. As we can expect, most of the text information will lose if the input text is rather long. In order to enable the model to be trained using long texts without changing the structure, we propose a multi-sentence weighting strategy.

\begin{figure}
	\centering
	\includegraphics[width=.35\textwidth]{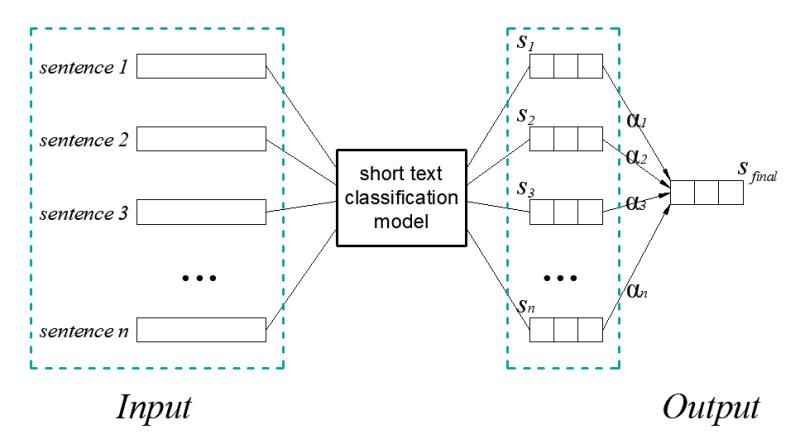}
	\caption{Multi-sentence classification. We ignore the order and association between sentences, taking the weighted average of sentences as the total result.}
	\label{weighting}
\end{figure}

As shown in Figure \ref{weighting}, we divide any input long text into $n_s$ parts. Generally speaking, it is wise to divide by sentences, denoted as $Input = \{sen_1, sen_2, ..., sen_{n_s}\}$. We apply the short text classification model to each part independently and obtain $n$ score distribution $\{\bm{p}_1, \bm{p}_2, ..., \bm{p}_{n_s}\}$. In order to simplify the question, we ignore the order relationships and associations of the text parts. The output of all the sentences $p$, is computed as a weighted sum of these distributions:
\begin{equation}
\bm{p}=\sum^{n_s}_{i=1}{\alpha_i \bm{p}_i}
\end{equation}
There are plenty of methods to set the value of sentence weights $\alpha$. A naive approach is to give the same weight to all sentences. \cite{Lin2016NeuralRE} use a selective attention mechanism to de-emphasize the noisy sentence. They use a matrix to evaluate the degree to which the input sentence matches a particular category as a weighted value. 

To simplify the interpretation process, we employ a method that do not add extra parameters to obtain the value of weights for sentences. We suppose that the higher the score of a sentence on a particular category is, the more important it is for the classification task. Conversely, an ambiguous sentence will receive low weight value. Inspired by the assumption, we use the maximum value of the score distribution $s$ as the weight value of the sentence. Specifically, we obtain the weight of each sentence by executing the softmax function on the maximum value of the score distribution obtained in different sentences. It is defined as:
\begin{equation}
\alpha_k = \frac{exp(max([\bm{s}_k]))}{\sum^{n_s}_{i}{exp(max([\bm{s}_i]))}}
\end{equation}


\section{Experiment}
 In this part, we design experiments to evaluate the performance and interpretability of the proposed model respectively. The experiments are performed on both Chinese and English datasets to prove that the proposed methods can be used in both languages.
\subsection{Comparison with State-of-the-art Classifiers}
\label{English_experiment}
We compare the classification performance of the proposed model with several state-of-the-art classifiers using different neural structures. The baselines include:

 \begin{itemize}
 	\item \textbf{n-grams + LR}: We collect some of the most frequent n-grams (5 times of the size of the training set) in the training set as the features and employ a LR to make the classification. 
 	\item \textbf{CNN}: The model proposed in \cite{Kim2014ConvolutionalNN} with pre-trained word embeddings. The most difference compared to our methods is the employment method of the pooling layer.
 	\item \textbf{Bi-lstm}: Bi-directional LSTM with pre-trained word embeddings.
 	\item \textbf{FastText}: The model proposed in \cite{joulin2016bag} which averages the word embeddings as the sentence representation.
 \end{itemize}

The datasets we use are described below: 

\begin{itemize}
	\item \textbf{Reuters\footnote{https://www.cs.umb.edu/~smimarog/textmining/datasets/}}: Reuters is a text classification dataset containing 21,578 samples. We use the same setting with \cite{yao2018graph} to split a R8 and a R52 version. 
	\item \textbf{TREC\footnote{http://cogcomp.org/Data/QA/QC/}}: TREC question dataset involves classifying a question into 6 question types (\emph{location}, \emph{entity}, \emph{abbreviation}, \emph{description}, \emph{human}, and \emph{number}) which is proposed in \cite{li2002learning}.  
	\item \textbf{Ohsumed\footnote{http://disi.unitn.it/moschitti/corpora.htm}}: The Ohsumed corpus is a bibiographic dataset containing 3,357 documents for training and 4,043 documents for testing. There are 23 target categories in total.
	\item \textbf{20NG\footnote{http://qwone.com/~jason/20Newsgroups/}}: Document classification dataset consists of 11,314 training samples and 7,532 test samples distributed over 20 target categories.
\end{itemize}


Datasets Ohsumed and 20Ng contain more than 1 sentence in each sample. We adopt two processing modes of treating all sentences as a whole and using mul-sentence weighting strategy for these two datasets. The proposed model uses randomly initialized word vectors. 
The embedding dimension $d_e$ and the feature vector dimension $d_m$ are set to 50. The maximum convolution length $l$ is set to 6.
For baseline models, some of the results refer to \cite{yao2018graph} and \cite{Kim2014ConvolutionalNN}, the others is reproduced with the default parameters in the original papers. The models where pre-trained vectors are needed employ 300-dimensional word vectors using the architecture of \cite{mikolov2013distributed}. 

 \begin{table}[!htbp]
	\centering
	\caption{Accuracy[\%] on classification tasks. \emph{out methods*} divide the input text into sentences and employs the proposed \emph{multi-sentence weighting strategy} for training.}
	\begin{tabular}{cccccc}
		\hline
		Model & R8 & R52 & TREC & Obsumed & 20NG\\
		\hline
		n-grams + LR & 93.7& 87.0& 87.2& 54.7& 83.2\\
		\hline
		CNN & 95.7& 87.6 & \textbf{91.2} & 58.4 &82.1\\
		\hline
		BI-lstm & 96.3 & 90.5 &90.6& 51.1 & 73.2\\
		\hline
		fastText & 96.1 & 92.8 &84.5&57.7 &79.4\\ 
		\hline
		our method & \textbf{97.0}& \textbf{93.1}&89.5& 58.7 & 83.4\\
		\hline
		our method* & - &-&-&\textbf{60.6}& \textbf{85.6}\\
		\hline
	\end{tabular}
	\label{benchmark}
\end{table}
We present the results in Table \ref{benchmark}. From the results we can see that our methods achieve great performance in most of the datasets. Our methods show poor performance on the dataset TREC, this may be because the scale of TREC is small and our method does not use the pre-trained word vectors. In Obsumed and 20NG, the \emph{multi-sentence weighting strategy} effectively promotes the performance of the model.




\subsection{Chinese Classification Task}
\label{Chinese task}
We also evaluate the performance of the model on Chinese tasks. The dataset we employed consists of some \emph{user survey feedback in taxi industry}. The target of the classification task is to distinguish the scope of the description. There are 7 target categories, including \emph{driver related}, \emph{products and services}, etc. There are a total of 13,525 instances in the dataset, and the distribution between classification categories is approximately equal. Each instance in the dataset consists of a title and a context of variable length. The length of the context varies greatly, ranging from 1 sentence to more than 100 sentences. 10\% of total instances are randomly split for testing. 

The experiment consisted of two groups. In the first group, the training process only uses the titles. And the second group use both of the title and the context. The baselines and the parameter settings are similar to Section \ref{English_experiment}.

\begin{table}[!htbp]
\centering
\caption{Accuracy [\%]) of different methods on taxi industry user feedback classification task. Our methods* employ \emph{multi-sentence weighting strategy}. \emph{Fasttext} and Bi-lstm(word) employ Chinese word-level models and the others use character-level models.}
\begin{tabular}{c|c|c}
	\hline
	 methods & title & title and context\\
	\hline
	fasttext & 75.4 & 79.0 \\
	\hline
	Bi-lstm (word) & 76.0 &  79.4\\
	\hline
	Bi-lstm (char) & 74.2 &78.3 \\
	\hline
	CNN & 81.8 &84.5 \\
	\hline
	our methods & 82.5 & 85.1\\
	\hline
	our methods* & - & \textbf{86.2}\\
	\hline
\end{tabular}
\end{table}

The results show that the proposed model outperforms other methods. Our methods with \emph{multi-sentence weighting strategy} achieve the best results due to the integration of title and context information.

\section{Analysis and Visualization}
 In this section, we analyze the interpretability of the proposed method on both English and Chinese datasets. The performance on English is assessed on the TREC dataset and the performance on Chinese is assessed on the proposed Chinese user survey feedback dataset.
 
\subsection{Interpreting Predictions with input tokens}

\begin{figure}
	\centering
	\includegraphics[width=.4\textwidth]{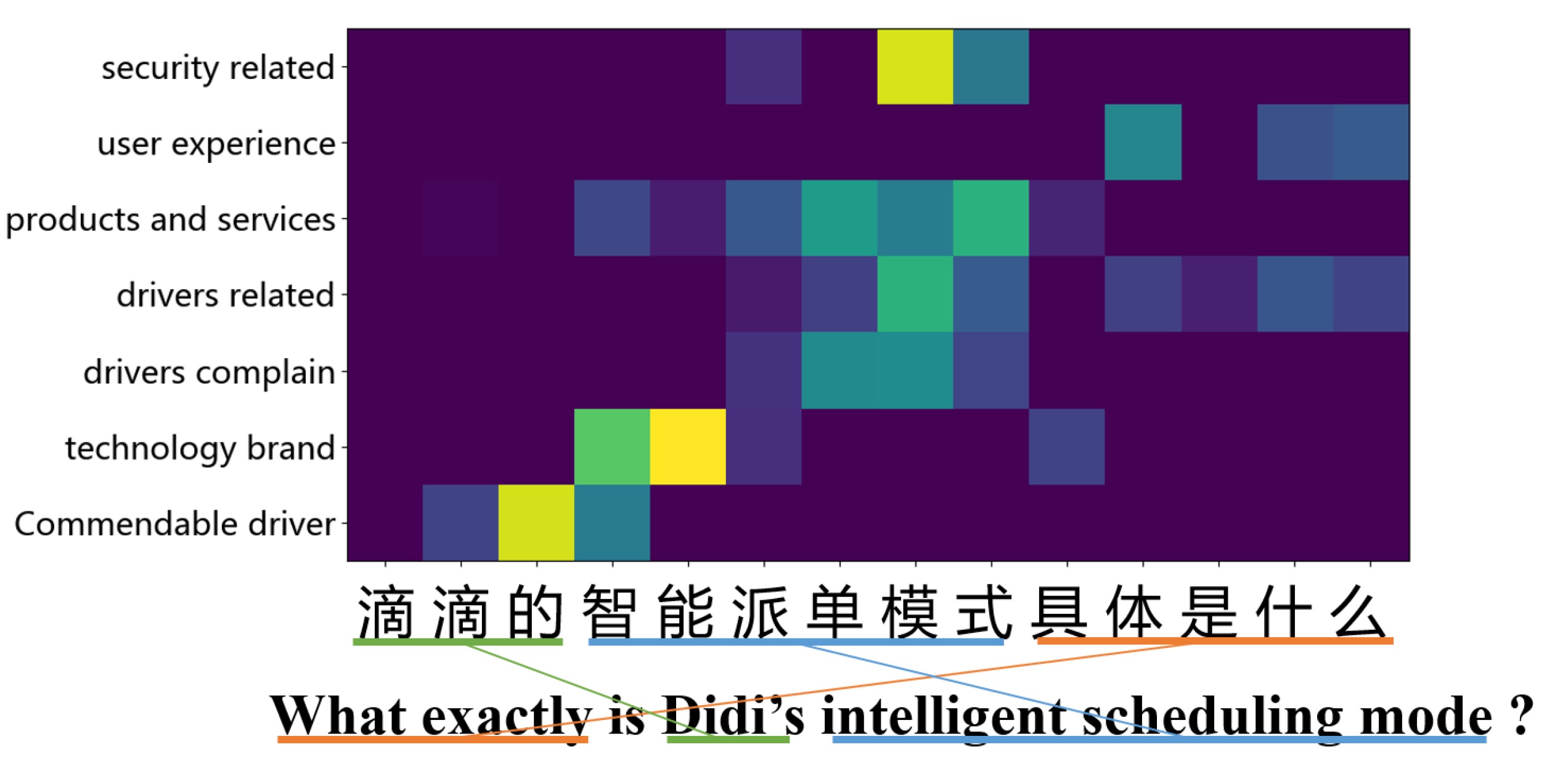}
	\caption{Visualizing the contribution for each character of the sentence on target categories. (\emph{Didi} is a taxi company)}
	\label{classification_interpretable}
\end{figure}
 
 \begin{figure}
 	\centering
 	\includegraphics[width=.4\textwidth]{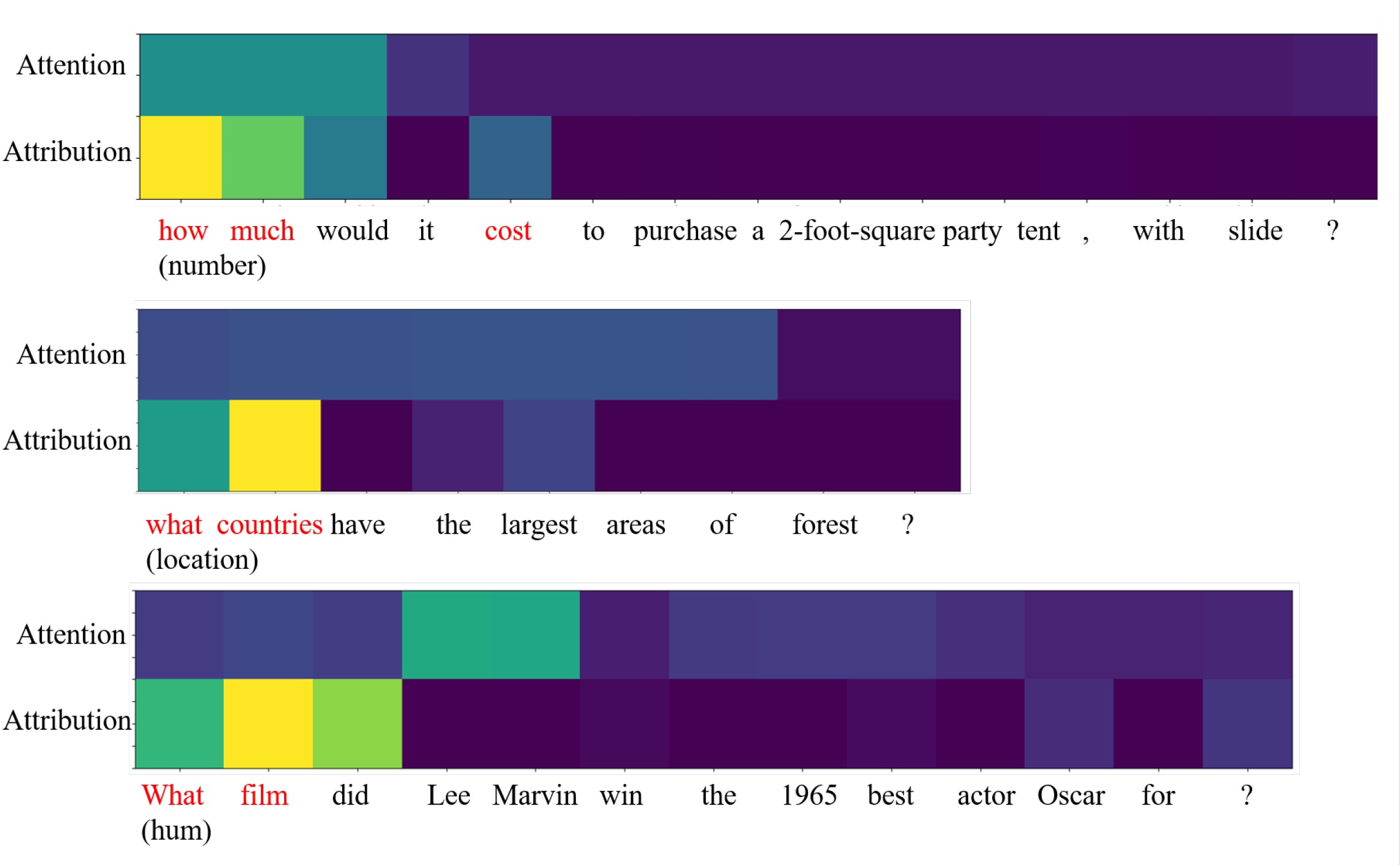}
 	\caption{Comparison of the interpreting result of the methods on TREC samples. \emph{Attention} visualize the weight of the attention layer for input tokens. \emph{Attribution} demonstrate the weight of input tokens on the ground category in our model.}
 	\label{attention}
 \end{figure}
 The interpretability of the proposed model is implemented by providing human-readable explanations for neural network predictions. After the proposed model converges on the training set, we can use the proposed methods to generate the contribution matrix. We visualize the contributions of the input tokens to the final categories to observe that which tokens in the input sentence that indicate the final predictions. 
 
 We evaluate the interpretable performance on the TREC dataset. The dataset involves classifying a question into 6 types (location, entity, abbreviation, description, human, and number). The attention mechanism is used as a baseline. We employ the attention mechanism in a Bi-lstm structure and the weights are visualized to indicate the importance of the input tokens. The proposed model is employed on the dataset and we visualize the weights of each word on the ground category. The examples are shown in Figure ~\ref{attention}.
 
 As the Figure ~\ref{attention} shown, the attention mechanism evaluates the importance of each word for the classification task, but the results are rather vague. It is difficult to locate certain words for the attention mechanism. In contrary, our model accurately locates the words that are related to the ground category.

In the text classification task, there are cases where different components in a sentence are associated with different target categories. We can use the interpretable result to prove that the proposed model can capture these information. As shown in Figure \ref{classification_interpretable}, \emph{intelligent} has a strong association with target category \emph{technology brand}. In addition, \emph{intelligent scheduling model} contributes to \emph{products and services}, \emph{scheduling model} contributes to \emph{driver related}. Most of the associations that the model has learned are reasonable.



\subsection{Interpreting Predictions with Samples}
\begin{figure}
	\centering
	\includegraphics[width=.46\textwidth]{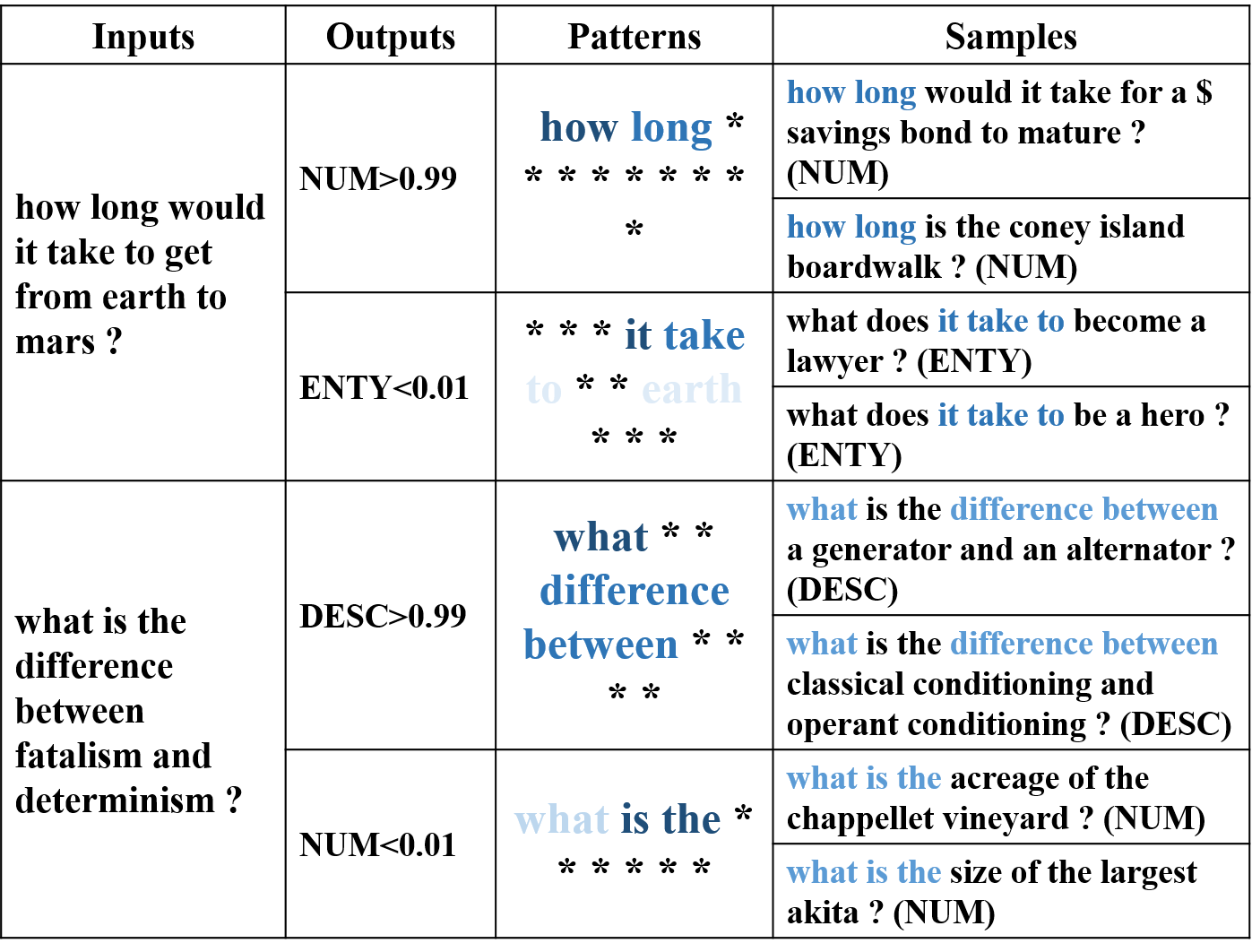}
	\caption{Example of interpreting predictions with samples. The samples come from \textbf{TREC} in Section \ref{English_experiment}. * in interpretable result denotes words with negative values. The deeper the word, the greater the value of the word for the corresponding category.}
	\label{interpretable_result}
\end{figure}
In text classification tasks, the effect of classification model depends heavily on the dataset. Limited by the scale of the dataset, the judgment bases of the model often differ from human cognition. In this section, we try to find out the judgment bases of the model in a specific classification prediction and further reason the bases with samples in the training set.

Given a input context, after the model make the prediction, we employ the proposed interpretable method to find out the value of each word in the context on target categories. We focus on the position of the words which have positive values on a particular category to generate a combination pattern. In general, the combination patterns can answer for what features indicate the model to make the prediction. Next, we look for the samples which have the highest suitability with the generated patterns in the training set. These samples usually answer how does the model learn the features.

Examples are shown in Figure \ref{interpretable_result}. The judgment of the model on dataset \textbf{TREC} is actually based on some combination patterns. For the first example, the model classifies the input sentence into \emph{NUM} with the confidence greater than 99\% solely based on the strong pattern \emph{how long}. As for the predictions with weak confidence, we can observe that most of the judgment bases are unreasonable combination patterns, such \emph{it take to} leads to \emph{ENTY} and \emph{what is the} leads to \emph{NUM}. We can figure out which samples of training set mislead the model to learn wrong features with the proposed method.

This work can be used to attribute the results of model mispredictions. By reasoning the error cases, we can find out the samples that mislead the model to learn wrong judgment bases. In this way, we can make directional adjustments to the judgments bases learned by the model.   

\section{Conclusion}

In this work, we propose a CNN text classification model and employ strategies including \emph{n-gram feature analysis} and \emph{convolution attribution} to interpreting the classification process. \emph{$N$-gram feature analysis} analyze the model prediction with the contributions of n-gram feature and \emph{convolution attribution} further establishes the relevance between n-gram features and word vectors. We also employ a \emph{multi-sentence weighting strategy} to employ the model in long text situation. Experiments show that the model has a similar performance to state-of-the-art classifiers and the multi-sentence weighting strategy works for long context input. Two applications with several interpretable cases are introduced to justify the interpretable performance of the model. In the future, we will further explore more methods to interpret neural models in NLP tasks and study how to use the interpretability of the model to directionally adjust model performance.

\clearpage
\bibliographystyle{named}
\bibliography{ijcai19}

\end{document}